\documentclass[a4paper]{article}

\usepackage{INTERSPEECH2021}

\usepackage{times}
\usepackage{url}
\usepackage{latexsym}

\usepackage[T1]{fontenc}

\usepackage[utf8]{inputenc}

\usepackage{xcolor}
\usepackage{graphicx}

\usepackage{amsmath}
\usepackage{bbm}

\usepackage{multirow}

\title{DEXTER: Deep Encoding of External Knowledge for Named Entity Recognition in Virtual Assistants}

\name{
 Deepak Muralidharan$^{1*}$\thanks{ $^*$\text{Equal contribution.}},
 Joel Ruben Antony Moniz$^{1*}$, Weicheng Zhang$^1$, Stephen Pulman$^2$,
 Lin Li$^1$, Megan Barnes$^{3\dagger}$\thanks{ $^\dagger$\text{Work done while at Apple.}}, Jingjing Pan$^1$,
 Jason Williams$^1$, Alex Acero$^1$}
\address{
  $^1$Apple, USA\\
   $^2$Apple, UK\\
  $^3$University of Washington, USA}
\email{deepak\_muralidharan@apple.com, joelrubenantony\_moniz@apple.com, weicheng\_zhang@apple.com,
spulman@apple.com, lli9@apple.com, mrbarnes@uw.edu, jpan9@apple.com, jason\_williams4@apple.com, aacero@apple.com}

\begin{document}
\maketitle
\begin{abstract}
	Named entity recognition (NER) is usually developed and tested on text
from well-written sources. However, in intelligent voice assistants, where NER is an important
component, input to NER may be noisy because of user or speech recognition error. In applications, entity labels may change frequently, and
non-textual properties like topicality or popularity
may be needed to choose among alternatives.

We describe a NER system intended to address these problems.
We test and train this system on a proprietary user-derived dataset. We compare with a baseline text-only NER system; the
baseline enhanced with external gazetteers; and the baseline enhanced with the search and indirect labelling
techniques we describe below. The final configuration gives around 6\% reduction in NER error rate.
We also show that this technique improves related tasks, such as semantic parsing, with an improvement of up to
5\% in error rate.

\end{abstract}

\section{Introduction}
\label{sec:intro}

NER is the process of labelling sequences of tokens in a  sentence with a label representing some
kind of semantic classification. Early NER approaches frequently relied on the use of patterns, rules and hand-crafted features for the identification of
named entities \cite{ratinov-roth-2009-design,finkel-manning-2009-joint,ritter-etal-2011-named}, often supplemented with
information from gazetteers. 
More recent approaches aim to learn to identify named entities and their types in a data-derived way \cite{passos-etal-2014-lexicon,ma-hovy-2016-end,lample-etal-2016-neural,chiu-nichols-2016-named}.
Typical NER systems are trained and tested on benchmark corpora such as CoNLL-2003
\cite{tjong-kim-sang-de-meulder-2003-introduction} or OntoNotes \cite{hovy-etal-2006-ontonotes,OntoNotesRelease5},
which are collections of documents derived from well-written sources. Such data are not representative of the
language encountered in many practical applications that require NER as a component, such as natural language
understanding in intelligent voice assistants, which is our focus here.

Firstly, for spoken language queries in a voice assistant, disambiguating between different entity
classes purely based on semantic word representations and the context of the utterance is difficult.
For example, in an utterance ``Play Bohemian Rhapsody'', the entity ``Bohemian Rhapsody'' is both a popular song as well
as a popular movie, and both entity classes are often referred to in the same context.
Non-linguistic contextual information, such as temporal popularity, relevance of the entity to the current application,
and other factors, are crucial for the NER model to effectively
predict the correct entity class.

Secondly, data-driven methods, which rely heavily on patterns available in the training data without additional
features, might not be able to handle less popular entities and templates with insufficient representation
in the training data.
For example, for an utterance like ``Play One by U2'', a purely data-driven model with only word embeddings might be
biased to predict the token ``one'' as non-entity text, whereas the user might actually be referring to the song called ``One'' by
``U2''.
(We have given the appropriate casing to this example, but an ASR system may not provide casing information at all).

Thirdly,  models trained on data collected from the past are usually unable to recognize newly released films,
songs, or albums. Time is always needed to collect and label enough training data to update models.

Finally, input to voice assistants is often noisy in various ways: speakers make errors and sometimes do not produce grammatically correct
or complete sentences, and ASR systems often mis-recognize what has been said.

In this paper,  we propose a novel method of encoding external knowledge about entities as part of NER model training.
We show that by treating entities as objects with attributes (such as popularity, associated
  textual strings,  entity class label) and encoding these entities through a neural network, the model learns correlations between the input tokens,
the attributes of entities  that these tokens
belong to, and their true class labels. We design a search engine trained on spoken queries to provide these entity candidates, which makes the
  system more resilient to the speech errors and user variation that commonly occur in spoken dialog systems. 
We show that using this approach, we are able to perform significantly better than an NER model trained only
using word and character embeddings, or models trained with additional implementations of gazetteer features.
As well as improvements on NER, we also show that our technique can easily be extended to other language
understanding tasks, such as semantic parsing, where we also see significant improvements over the baseline.

\section{Related Work}
\label{sec:rel_work}

External knowledge sources  to improve NER task performance has been explored in previous literature.
One approach has been the use of Wikipedia articles to synthetically generate training data.
\cite{ghaddar-langlais-2017-winer} classified Wikipedia articles into named entity types and then generated silver-standard training annotations
for NER by transforming links between Wikipedia articles into named entity annotations by projecting the target
article's classifications onto the anchor text.
\cite{multilingual-ner} used Wikipedia to automatically generate additional annotated data for training NER systems by
extending annotations for non anchored strings using coreference information.

However, the most common approach has been through the use of gazetteer features.
\cite{chiu-nichols-2016-named} proposed a new lexicon encoding scheme and matching algorithm by constructing uni-grams and bi-grams
to make use of partial matches and combined the encoding scheme with a hybrid LSTM-CNN based neural NER model.
\cite{liu-etal-2019-towards} included gazetteer features in a NER model trained using
hybrid semi-Markov conditional random fields (HSCRFs) by introducing an additional module that
scores a candidate entity span by the degree to which the entity span softly matches the gazetteer.
\cite{peshterliev2020selfattention} enhanced gazetteer matching with multi-token and single-token matches in the same representation and
combined the matching with a self-attention model to generate gazetteer embeddings.

In our approach, we do not use gazetteers directly, but harvest similar information from a knowledge
graph. However, these token sequences are not the only properties associated with entities. Furthermore, unlike
\cite{peshterliev2020selfattention,liu-etal-2019-towards}, we do not assume that exact matching is either necessary or
sufficient to be able to use this information. As described above, in a spoken dialog setting we cannot assume that
input will be complete or correct. It is of course possible to  devise complex encoding schemes for gazetteers to support partial
matches,
as in \cite{chiu-nichols-2016-named}. However, we instead use the knowledge graph to construct a search engine tuned to spoken queries to retrieve a ranked list of entity candidates.
This approach enables our system to be robust to partial matches,
speech errors, and other variations in named entities observed in usage.

\section{Methodology and Design}
\label{sec:method}

In this section, we will explain the architecture of our NER system (as shown in Appendix Figure \ref{fig:dense}).

\label{subsec:external-resources}

\noindent
{\bf Incorporating External Resources:} We conceptualize entities as objects that have various document properties associated with them: an entity class
label, in particular. Entities that can have different class labels are different entities.
To construct the properties associated with entities,
we first crawl a knowledge graph, which is updated daily.  This crawl gives us our
universe of entities. From the text descriptions associated with nodes, and the relations they take part in (as indicated by incoming and
outgoing edges in the graph), we construct n-grams of normalized (lower cased, stemmed, and with punctuation removed) tokens that may be used to refer to the entity. We
augment these n-grams with aliases and acronyms mined from various sources. For example, the entity
representing ``San Francisco International Airport'' will also be associated with ``SFO'', ``San Fran Airport'', etc. We then create an inverted index from these n-grams to the entities, and compute and store the tf-idf
scores for all n-grams: tf is computed over the concatenation of all
  n-grams for an entity and idf is computed over the set of n-grams for all entities.  We also compute and store a popularity score for each entity. This score is the prior probability based on frequency of occurrence in
anonymized usage logs and engagement with first party apps.

We now have a list of around 10m entities with their associated properties and can efficiently search for
candidate entities given an n-gram of tokens. We make this search more robust to ASR and user errors by training a ``correction model'' learned
from common usage patterns. For example, ``Cincinnati Bangles'' would be corrected to ``Cincinnati Bengals'', and
``Conor Gregor'' would be corrected to ``Conor McGregor''.

Note that the set of entity class labels derived from the knowledge graph is not identical to the final set of NE labels
assigned by the network described in Section \ref{subsec:dense_aggregation}. There are 32 entity classes in the knowledge graph, and the
network learns to map these to 23 NER labels. For example, the fine-grained entities ``city'', ``state'', ``country''
are all mapped to ``location''. Some NER labels, like ``dateTime'', in fact do not correspond to anything in the knowledge graph,
and the network learns to assign this on the basis of properties of tokens in context. In effect, the knowledge graph
entity classes serve as features for the network's assignment of NER labels.

\label{subsec:dense_aggregation}
\noindent
{\bf Dense Aggregation Layer:} For a given input utterance, we tokenize it and construct all the n-grams that
each token is part of, up to a specific window length. For example, a token is a member of one uni-gram, potentially two
bi-grams, three tri-grams etc.

For each n-gram that the token is part of, we issue a query into the search engine, to get candidate
lists of entities. We then use a simple linear scorer which uses features such as tf-idf score and popularity to generate a
ranked list of these entities.
From this ranked list, we select the top-$k$ entities for each of the entity classes in our ontology to get a
varied distribution of entities across all entity classes for every n-gram.

For every token in the utterance, we now have a list of entity candidates retrieved from the search engine for all the n-grams that the
token is part of, up to a window size $w$.
To get the cumulative importance of an entity for a token across its n-gram context, we concatenate the tf-idf scores
of the entity across all the n-grams indexing it,\footnote[1]{if an entity is not retrieved as part of an n-gram, we simply give it
  a tf-idf score of 0} along with the popularity score of the entity.
We then feed this vector through a single layer perceptron to obtain a score for the entity with respect to this
token and its n-gram context. We do this for all the entities retrieved for that token.

More formally, for a token $t$ and a retrieved entity candidate $e_{i}$ (where $i$ corresponds to the index of the
entity in the complete listing of entities $E$),
and a window size $w$, the tf-idf scores of $e_{i}$ are considered with respect to all uni-, bi-, tri-,\dots,$w$-grams that contain $t$.
Let these scores be represented by $r_{1,1}$ for the unigram score; $r_{2,1}$ and $r_{2,2}$ for the two bigram scores; \ $r_{w,1}\ldots r_{w,w}$ for the $w$ $w$-gram scores.
Let $p$ be the popularity score of the entity and $c$ represent the class of the entity as an integer. Then the vector
created by the concatenation of tf-idf scores with the popularity score is:
\begin{align}
  \mathbf{r} &= [r_{1,1};r_{2,1};r_{2,2};\ldots r_{w,1}; \ldots;r_{w,w}; p].
\end{align}
Let $o(.)$ represent a function that converts an integer into a one-hot vector with as many dimensions are there are
entity classes.
We first multiply $\mathbf{r}$ by this one-hot vector: $o(c)^T \cdot \mathbf{r} $. This gives us a matrix where one row contains the
values from $\mathbf{r}$, with zeros elsewhere. We reshape the matrix to a vector $v$ and pass $v$ through a linear layer with activation function
$a(.)$, weight vector $\mathbf{w}$, and bias $\mathbf{b}$, yielding a single scalar as output. The parameters of the
linear layer are shared across all the entities.
\begin{align}
     s_{e_{i}} &= a(\mathbf{w} \cdot v + \mathbf{b}).
\end{align}
\label{eqn:docscore}
Once we have computed the scores for each entity retrieved for each token $t$, we create a vector for each token of
length
 $|C|$, where $C$ is the set of entity classes supported
in our ontology. Then we put in the corresponding position of the vector the highest score that any entity of that class
got for that token in that n-gram context, denoted by $\mathbf{u}_t$.
This operation is referred to as the ``Reduction layer" block in Figure \ref{fig:dense} (a).

\label{subsec:contextual_cnn}
\noindent
{\bf Capturing Contextual Entity Signals:} 
To encourage the model to use contextual information, we also incorporate a layer of 1D CNNs on top of the dense
aggregation layer. This layer provides several benefits. Firstly, the layer
allows each token to have a contextual representation that includes features from neighbouring tokens. 
Secondly, this CNN layer can learn from complex dependencies present between tokens other than the current token, and
across the different entity classes; for example,
the presence of a music album helps disambiguate between the name of a music artist or a movie actor.
Thirdly, the CNN adds two useful inductive biases: location invariance and edge detection \cite{Lecun98gradient-basedlearning,kim-2014-convolutional}. The location invariance ensures that features corresponding to similar utterances (such as song titles) are treated similarly irrespective of where they are present in the utterance. The ability of the correlation operation to detect edges enables the featurization to be biased to favor entities consistently retrieved across tokens for a certain entity class since a match over longer spans within an utterance is likely to be better than if each token were to match to a different entity.

The input to the convolution layer $f_{conv}(.)$ is the output of the dense aggregation layer,
with shape of $T\times|C|$. 
In this approach, we convolve with a window size of $w$ (we set $w=7$) centered at each token in the $T$ dimension. Let the second dimension, of size $|C|$, represent the input channels. We set the number of convolutional filters (i.e., output channels) to be 32.
The output representations
\begin{equation}
    \mathbf{q}_t = f_{conv}(\mathbf{u}_{t-\frac{w-1}{2}}\cdots\mathbf{u}_{t+\frac{w-1}{2}}),
\end{equation}
will be used in the following tasks as the DEXTER embedding.

\label{subsec:nertraining}
\noindent
{\bf NER Training:} For training our baseline NER model, we use an architecture similar to \cite{lample-etal-2016-neural} and \cite{chiu-nichols-2016-named}
where for every token, character-level features are extracted by a bi-LSTM and concatenated with pre-trained
GLoVe embeddings \cite{pennington-etal-2014-glove}. This character-word representation is passed through a sequence
level bi-LSTM and fed into a CRF model to produce the final labels.

For our extended system,  we concatenate our DEXTER embeddings produced as described in the previous section to these word and character embeddings
before feeding the combined embeddings to the upper layers. The weights that contribute towards the DEXTER embeddings are learned end-to-end
during training, without pre-training, additional loss signals, or supplementary training data.
The architecture is shown in Figure \ref{fig:dense}(c).

\label{subsec:sptraining}
\noindent
{\bf Shallow Semantic Parsing:}
We also measure DEXTER's impact on
shallow semantic parsing, which involves jointly identifying the user's intent and
assigning a semantic label to each word in an utterance. For example, in ``Play the movie Moana'',
the model identifies the user intent as ``play'', tags ``movie'' as the type of entity to be played, and tags ``Moana''
as the name of the movie. Unlike NER, the semantic parsing task is domain-specific, with each domain having a different set of
intents and semantic labels.  Furthermore, we are not looking for units
above the word level:  the semantic labels are assigned to every word in the utterance and are not only for named entities.

For this task, we compare against a model that uses a bi-LSTM with word embeddings and gazetteer features as input (similar to \cite{muralidharan2021noise}), with our baseline showing performance without gazetteer features. We compare these settings with a version replacing the gazetteer with DEXTER, and show performance across 11 domains in Section \ref{subsec:sem_parse}. As in the case of our NER setup, we train the DEXTER embeddings and the model end-to-end.

\section{Datasets and Training Methodology}
\label{sec:expt_res}

There is no public benchmark dataset that we are aware of that has the properties we need, and so we
developed our own in-house one.
We randomly sampled around $500k$ utterances belonging to entity rich domains like music and sports
from usage logs, then anonymized and annotated them for NE labels.
For the annotation, we followed a standard B-I-O tagging approach for NE labels designed specifically for our ontology,
which consists of 23 fine-grained entity classes such as ``song'', ``artist'', ``celebrity'', ``athlete'' etc.
We use 70\% of the graded data as our training set, 20\% as development set and 10\% data as a held-out test set.
To evaluate the NER model performance, we report results on the ``music'', ``sports'' and ``movies \& TV'' domains, and
use the standard NER-F1 metric used for the CoNLL-2003 shared task \cite{tjong-kim-sang-de-meulder-2003-introduction}.
In our data, the average utterance length is 5.03 tokens. 32.30\% of utterances contain no entities, 55.44\% have 1 entity and 11.57\% have 2 entities. 81.66\% of all entities can be linked to the knowledge graph, of which ~84\% exactly match their canonical name, while the rest contain a user error (~9\%) or an ASR error (~7\%).

We train the NER model using a standard mini-batch gradient descent with a batch size of 128 and using the Adam optimizer \cite{kingma2014adam} with an initial
learning rate of 0.001. We evaluate the validation accuracy every 1000 training iterations, and decay the learning rate by a factor of 0.9 when the validation F1 score fails to improve by more than 1e-5. We train the NER model until the learning rate falls below 1e-7, for at most 50 epochs in total. We apply a dropout \cite{srivastava2014dropout} of 0.6 to the input of the 
bi-directional LSTM layer to prevent over-fitting.
The word embeddings and character embeddings both have 200 dimensions. The bidirectional-LSTM encoding the characters has a hidden state size of 100 dimensions. The bidirectional-LSTM whose output feeds into the CRF has 450 hidden state dimensions.

In the DEXTER embeddings module, we set the n-gram context window for tokens to 3 (uni-grams, bi-grams and tri-grams) and
retrieve top-10 candidates per entity class for every n-gram search query issued to the search engine.
In the CNN layer used to capture contextual similarity, we use 32 convolutional kernels of width 7. We apply padding of size 3 on either side, to keep the output dimension the same as that of the input. A tanh non-linearity is applied after the CNN layer.

\section{Results}
\label{sec:results}

\subsection{NER}
\label{subsec:res_ner}

\begin{table}[h]
  \caption{\text{*} indicates that the results improve statistically significantly ($p<0.05$) over the (a)
    baseline, with the $p$-values shown in parentheses, calculated as the one-tailed p-value comparing the binomial distributions formed by assuming the utterance-level correctness of a setting as binomial variable. Note that all numbers reported are over 10 runs.}
    \label{tb:ner-result}
\begin{tabular}{|l|l|l|l|}
\hline \bf & movies \& TV & music & sports\\ \hline
a & 79.26  & 83.51  & 91.61 \\
b & 79.43 & 83.69 & 91.66 \\
c & 79.43  & 84.14  & 91.66  \\
d & 80.08 & 84.70 & 91.70 \\
e  & \textbf{80.48}* {(5.5e-7)}  & \textbf{85.10}*{ (1.1e-8)}  & \textbf{91.74}{ (0.098)} \\
\hline
\end{tabular}
\end{table}

We compare the NER-F1 scores across five settings, taking existing or new implementations of comparable approaches:
(a) a baseline for which we use the implementation from \cite{lample-etal-2016-neural} which uses only word and character embeddings,
(b) a system implemented similarly to \cite{liu-etal-2019-towards} which uses a gazetteer-enhanced sub-tagger (using token embeddings in our case) and performs soft dictionary matching to generate gazetteer embeddings
which are concatenated to the input word and character embeddings in (a),
(c) a system with token matched gazetteer features implemented according to \cite{ratinov-roth-2009-design} concatenated to the input word and character embeddings in (a),
(d) a system implemented according to \cite{peshterliev2020selfattention} which uses self-attention and match span encoding to build enhanced gazetteer embeddings
which are concatenated to the input word and character embeddings in (a) and finally, (e) our proposed approach using DEXTER embeddings concatenated to the input word and character embeddings in (a).
All of the models are trained on $500k$ training data equally distributed amongst the three domains: music, movies \& TV and sports.
The gazetteers used in (b), (c) and (d) are constructed from a flattened version of the same knowledge graph used in (e) to keep the entity information consistent across all settings.
The F1 scores are reported on the held-out test sets for the best model selected by our development sets.

From Table \ref{tb:ner-result}, we see that external knowledge is essential to
get good accuracies in a fine-grained entity class setting. 
The models which use soft gazetteer features \cite{liu-etal-2019-towards} and token-match gazetteer features \cite{ratinov-roth-2009-design} do relatively
better than the baseline implementation of \cite{lample-etal-2016-neural}, which uses only word and character embeddings.
The model which uses self-attention and multi-token match span encoding \cite{peshterliev2020selfattention}
to generate gazetteer embeddings does better still.
Finally, our proposed approach which uses DEXTER embeddings further improves over all of the above described approaches and
shows a statistically significant ($p<0.05$) F1 score improvement of around 1\% on average (5.7\% reduction in error rate) across all three test sets over \cite{lample-etal-2016-neural}.

\begin{table}[b]
  \centering
  \caption{Ablations
  }
  \label{tb:ablation-result}
\begin{tabular}{|l|c|c|c|}
\hline \bf & movie \& TV & music & sports \\ \hline
a &  \bf 80.48  & \bf 85.10  & \bf 91.74  \\
b &  79.36  &  83.89  &  91.59  \\
c  &  80.38  &  84.92  & \bf 91.74  \\
d &  80.32  &  84.87  &  91.65  \\
e  &  80.31  &  85.07  & 91.67  \\
\hline
\end{tabular}
\end{table}

In Table \ref{tb:ablation-result}, we perform an ablation study on different model settings
and show the individual impact of our architecture's components.
The complete system is (a) and  replacing (b) the single layer perceptron (SLP) (denoted by $\{\mathbf{w},\mathbf{b}\}$ in Equation \ref{eqn:docscore}) from the best model setting with a
max pooling aggregation leads to a drop in NER-F1 accuracy by around 0.83\% points (averaged across the 3 domains).
The single layer perceptron allows the model to learn the scaling across the tf-idf scores from
different n-gram searches which a simple max pool aggregation does not provide.
We also show (c) replacing the class-specific SLP with a class-agnostic SLP leads to
an accuracy drop of around 0.1\%.
This accuracy drop could be explained by the fact that the distribution of attributes such as popularity within an entity class
widely varies across entity classes, and the additional parameters in the class-specific model allow the system
to capture this difference and scale across different distributions. Further, (d) tying the CNN weights so that the
convolution operation performed uses the same set of convolution filters irrespective of the entity type causes the NER
F-1 to drop by 0.16\%. Finally, (e) when we remove the CNN layer from the network, the NER-F1 drops by 0.1 points
(on average across all three test sets) which shows that the CNN allows the
model to capture context across multiple tokens, thereby helping improve the NER accuracy.\\

\subsection{Shallow Semantic Parser}
\label{subsec:sem_parse}

In addition to showing the effectiveness of the DEXTER encoding on named entity recognition (Section
\ref{subsec:res_ner}), we also explore 
DEXTER in shallow semantic parsing, as described in Section \ref{subsec:sptraining}. Appendix Table \ref{tab:sp}, but we show that
a token-match gazetteer feature
approach improves over a model which uses only word and character embeddings, and 
that shallow semantic parser models trained with DEXTER significantly improve over models trained with
token-match gazetteer features in 5 out of 11 domains in our test sets. This indicates the superiority of our encoding scheme.
We see an improvement of around 5.4\% reduction in error rate (averaged across all the 11 domains) over the model which uses only word and character embeddings and an
improvement of around 3.1\% over the model which also uses token-match gazetteers.
In the domains that do not show statistically significant improvements even with
DEXTER (such as ``Photos'', ``Notebook'' and ``Clock''), we hypothesize that this is due to the low coverage of NE spans for these domains
in our knowledge graph, or because these domains have few named entities present.

\section{Conclusion}
\label{subsec:conclusion}
We conclude with illustrative examples.
DEXTER  improves NER in cases where the transcribed utterance is noisy, e.g. in  ``Find I know you did last summer'', where the user's intent is to
find the movie ``I know \textbf{what} you did last summer'', the utterance does not provide the exact name of
the entity and there is not enough information in the local context for the NER model to predict accurately.
However, with DEXTER, we are able to recover by retrieving entities even if
they only partially match, and by using attributes of the retrieved entity
(popularity of the movie and tf-idf) allowing the NER model to detect the NE label correctly as a ``movie''.

Secondly, DEXTER is able to improve NER in scenarios where the NE label is heavily
influenced by current interests and fashions.
For example, in ``Play Godzilla'', the NE ``Godzilla'' is a famous movie as well as a very popular song by the world-renowned artist Eminem, and there is not enough context for the model to disambiguate effectively.
However, in DEXTER, we retrieve entities belonging to multiple entity classes (such as ``movie'', ``song'', ``artist'') and the
trained single layer perceptron weights entities across different entity classes based on their attributes, such as freshness and popularity.
Since at the relevant time (when it was released) the ``movie'' label would have been more popular than the ``song'', the NER model was able to use this
discriminatory feature and predict the NE label accurately.
However, if the transcribed utterance was ``Play the song Godzilla'' or ``Play Godzilla by Eminem'',
our model respects the semantic context of the utterance (i.e., the user's request is for the song to be played, as opposed to the movie)
and predicts the ``song'' NE label even though the ``movie'' might have been more popular.

\bibliographystyle{IEEEtran}
\bibliography{biblio/biblio,biblio/biblio_deepak,biblio/biblio_joel,biblio/biblio_stephen}

\begin{thebibliography}{10}
\providecommand{\url}[1]{#1}
\csname url@samestyle\endcsname
\providecommand{\newblock}{\relax}
\providecommand{\bibinfo}[2]{#2}
\providecommand{\BIBentrySTDinterwordspacing}{\spaceskip=0pt\relax}
\providecommand{\BIBentryALTinterwordstretchfactor}{4}
\providecommand{\BIBentryALTinterwordspacing}{\spaceskip=\fontdimen2\font plus
\BIBentryALTinterwordstretchfactor\fontdimen3\font minus
  \fontdimen4\font\relax}
\providecommand{\BIBforeignlanguage}[2]{{%
\expandafter\ifx\csname l@#1\endcsname\relax
\typeout{** WARNING: IEEEtran.bst: No hyphenation pattern has been}%
\typeout{** loaded for the language `#1'. Using the pattern for}%
\typeout{** the default language instead.}%
\else
\language=\csname l@#1\endcsname
\fi
#2}}
\providecommand{\BIBdecl}{\relax}
\BIBdecl

\bibitem{ratinov-roth-2009-design}
\BIBentryALTinterwordspacing
L.~Ratinov and D.~Roth, ``Design challenges and misconceptions in named entity
  recognition,'' in \emph{Proceedings of the Thirteenth Conference on
  Computational Natural Language Learning ({C}o{NLL}-2009)}.\hskip 1em plus
  0.5em minus 0.4em\relax Boulder, Colorado: Association for Computational
  Linguistics, Jun. 2009, pp. 147--155. [Online]. Available:
  \url{https://www.aclweb.org/anthology/W09-1119}
\BIBentrySTDinterwordspacing

\bibitem{finkel-manning-2009-joint}
\BIBentryALTinterwordspacing
J.~R. Finkel and C.~D. Manning, ``Joint parsing and named entity recognition,''
  in \emph{Proceedings of Human Language Technologies: The 2009 Annual
  Conference of the North {A}merican Chapter of the Association for
  Computational Linguistics}.\hskip 1em plus 0.5em minus 0.4em\relax Boulder,
  Colorado: Association for Computational Linguistics, Jun. 2009, pp. 326--334.
  [Online]. Available: \url{https://www.aclweb.org/anthology/N09-1037}
\BIBentrySTDinterwordspacing

\bibitem{ritter-etal-2011-named}
\BIBentryALTinterwordspacing
A.~Ritter, S.~Clark, {Mausam}, and O.~Etzioni, ``Named entity recognition in
  tweets: An experimental study,'' in \emph{Proceedings of the 2011 Conference
  on Empirical Methods in Natural Language Processing}.\hskip 1em plus 0.5em
  minus 0.4em\relax Edinburgh, Scotland, UK.: Association for Computational
  Linguistics, Jul. 2011, pp. 1524--1534. [Online]. Available:
  \url{https://www.aclweb.org/anthology/D11-1141}
\BIBentrySTDinterwordspacing

\bibitem{passos-etal-2014-lexicon}
\BIBentryALTinterwordspacing
A.~Passos, V.~Kumar, and A.~McCallum, ``Lexicon infused phrase embeddings for
  named entity resolution,'' in \emph{Proceedings of the Eighteenth Conference
  on Computational Natural Language Learning}.\hskip 1em plus 0.5em minus
  0.4em\relax Ann Arbor, Michigan: Association for Computational Linguistics,
  Jun. 2014, pp. 78--86. [Online]. Available:
  \url{https://www.aclweb.org/anthology/W14-1609}
\BIBentrySTDinterwordspacing

\bibitem{ma-hovy-2016-end}
\BIBentryALTinterwordspacing
X.~Ma and E.~Hovy, ``End-to-end sequence labeling via bi-directional
  {LSTM}-{CNN}s-{CRF},'' in \emph{Proceedings of the 54th Annual Meeting of the
  Association for Computational Linguistics (Volume 1: Long Papers)}.\hskip 1em
  plus 0.5em minus 0.4em\relax Berlin, Germany: Association for Computational
  Linguistics, Aug. 2016, pp. 1064--1074. [Online]. Available:
  \url{https://www.aclweb.org/anthology/P16-1101}
\BIBentrySTDinterwordspacing

\bibitem{lample-etal-2016-neural}
\BIBentryALTinterwordspacing
G.~Lample, M.~Ballesteros, S.~Subramanian, K.~Kawakami, and C.~Dyer, ``Neural
  architectures for named entity recognition,'' in \emph{Proceedings of the
  2016 Conference of the North {A}merican Chapter of the Association for
  Computational Linguistics: Human Language Technologies}.\hskip 1em plus 0.5em
  minus 0.4em\relax San Diego, California: Association for Computational
  Linguistics, Jun. 2016, pp. 260--270. [Online]. Available:
  \url{https://www.aclweb.org/anthology/N16-1030}
\BIBentrySTDinterwordspacing

\bibitem{chiu-nichols-2016-named}
\BIBentryALTinterwordspacing
J.~P. Chiu and E.~Nichols, ``Named entity recognition with bidirectional
  {LSTM}-{CNN}s,'' \emph{Transactions of the Association for Computational
  Linguistics}, vol.~4, pp. 357--370, 2016. [Online]. Available:
  \url{https://www.aclweb.org/anthology/Q16-1026}
\BIBentrySTDinterwordspacing

\bibitem{tjong-kim-sang-de-meulder-2003-introduction}
\BIBentryALTinterwordspacing
E.~F. Tjong Kim~Sang and F.~De~Meulder, ``Introduction to the {C}o{NLL}-2003
  shared task: Language-independent named entity recognition,'' in
  \emph{Proceedings of the Seventh Conference on Natural Language Learning at
  {HLT}-{NAACL} 2003}, 2003, pp. 142--147. [Online]. Available:
  \url{https://www.aclweb.org/anthology/W03-0419}
\BIBentrySTDinterwordspacing

\bibitem{hovy-etal-2006-ontonotes}
\BIBentryALTinterwordspacing
E.~Hovy, M.~Marcus, M.~Palmer, L.~Ramshaw, and R.~Weischedel, ``{O}nto{N}otes:
  The 90{\%} solution,'' in \emph{Proceedings of the Human Language Technology
  Conference of the {NAACL}, Companion Volume: Short Papers}.\hskip 1em plus
  0.5em minus 0.4em\relax New York City, USA: Association for Computational
  Linguistics, Jun. 2006, pp. 57--60. [Online]. Available:
  \url{https://www.aclweb.org/anthology/N06-2015}
\BIBentrySTDinterwordspacing

\bibitem{OntoNotesRelease5}
\BIBentryALTinterwordspacing
R.~Weischedel, M.~Palmer, M.~Marcus, E.~Hovy, S.~Pradhan, L.~Ramshaw, N.~Xue,
  A.~Taylor, J.~Kaufman, M.~Franchini, M.~El-Bachouti, R.~Belvin, and
  A.~Houston, ``Ontonotes release 5.0,'' Linguistic Data Consortium, Tech.
  Rep., 2013. [Online]. Available:
  \url{https://catalog.ldc.upenn.edu/docs/LDC2013T19/OntoNotes-Release-5.0.pdf}
\BIBentrySTDinterwordspacing

\bibitem{ghaddar-langlais-2017-winer}
\BIBentryALTinterwordspacing
A.~Ghaddar and P.~Langlais, ``{W}i{NER}: A {W}ikipedia annotated corpus for
  named entity recognition,'' in \emph{Proceedings of the Eighth International
  Joint Conference on Natural Language Processing (Volume 1: Long
  Papers)}.\hskip 1em plus 0.5em minus 0.4em\relax Taipei, Taiwan: Asian
  Federation of Natural Language Processing, Nov. 2017, pp. 413--422. [Online].
  Available: \url{https://www.aclweb.org/anthology/I17-1042}
\BIBentrySTDinterwordspacing

\bibitem{multilingual-ner}
J.~Nothman, N.~Ringland, W.~Radford, T.~Murphy, and J.~Curran, ``Learning
  multilingual named entity recognition from wikipedia,'' in \emph{Artificial
  Intelligence}.\hskip 1em plus 0.5em minus 0.4em\relax Elsevier, Jan. 2013,
  pp. 151--175.

\bibitem{liu-etal-2019-towards}
\BIBentryALTinterwordspacing
T.~Liu, J.-G. Yao, and C.-Y. Lin, ``Towards improving neural named entity
  recognition with gazetteers,'' in \emph{Proceedings of the 57th Annual
  Meeting of the Association for Computational Linguistics}.\hskip 1em plus
  0.5em minus 0.4em\relax Florence, Italy: Association for Computational
  Linguistics, Jul. 2019, pp. 5301--5307. [Online]. Available:
  \url{https://www.aclweb.org/anthology/P19-1524}
\BIBentrySTDinterwordspacing

\bibitem{peshterliev2020selfattention}
S.~Peshterliev, C.~Dupuy, and I.~Kiss, ``Self-attention gazetteer embeddings
  for named-entity recognition,'' \emph{arXiv preprint arXiv:2004.04060}, 2020.

\bibitem{Lecun98gradient-basedlearning}
Y.~Lecun, L.~Bottou, Y.~Bengio, and P.~Haffner, ``Gradient-based learning
  applied to document recognition,'' in \emph{Proceedings of the IEEE}, 1998,
  pp. 2278--2324.

\bibitem{kim-2014-convolutional}
\BIBentryALTinterwordspacing
Y.~Kim, ``Convolutional neural networks for sentence classification,'' in
  \emph{Proceedings of the 2014 Conference on Empirical Methods in Natural
  Language Processing ({EMNLP})}.\hskip 1em plus 0.5em minus 0.4em\relax Doha,
  Qatar: Association for Computational Linguistics, Oct. 2014, pp. 1746--1751.
  [Online]. Available: \url{https://www.aclweb.org/anthology/D14-1181}
\BIBentrySTDinterwordspacing

\bibitem{pennington-etal-2014-glove}
\BIBentryALTinterwordspacing
J.~Pennington, R.~Socher, and C.~Manning, ``{G}lo{V}e: Global vectors for word
  representation,'' in \emph{Proceedings of the 2014 Conference on Empirical
  Methods in Natural Language Processing ({EMNLP})}.\hskip 1em plus 0.5em minus
  0.4em\relax Doha, Qatar: Association for Computational Linguistics, Oct.
  2014, pp. 1532--1543. [Online]. Available:
  \url{https://www.aclweb.org/anthology/D14-1162}
\BIBentrySTDinterwordspacing

\bibitem{muralidharan2021noise}
D.~Muralidharan, J.~R.~A. Moniz, S.~Gao, X.~Yang, J.~Kao, S.~Pulman,
  A.~Kothari, R.~Shen, Y.~Pan, V.~Kaul \emph{et~al.}, ``Noise robust named
  entity understanding for voice assistants,'' in \emph{Proceedings of the 2021
  Conference of the North American Chapter of the Association for Computational
  Linguistics: Human Language Technologies: Industry Papers}, 2021, pp.
  196--204.

\bibitem{kingma2014adam}
D.~P. Kingma and J.~Ba, ``Adam: A method for stochastic optimization,''
  \emph{arXiv preprint arXiv:1412.6980}, 2014.

\bibitem{srivastava2014dropout}
N.~Srivastava, G.~Hinton, A.~Krizhevsky, I.~Sutskever, and R.~Salakhutdinov,
  ``Dropout: a simple way to prevent neural networks from overfitting,''
  \emph{The journal of machine learning research}, vol.~15, no.~1, pp.
  1929--1958, 2014.

\end{thebibliography}

\appendix
\onecolumn

\section{Architecture Diagram}
\label{sec:app_sp_table}
\vspace{-0.23in}
\begin{figure}[h]
  \centering
  \includegraphics[width=\linewidth]{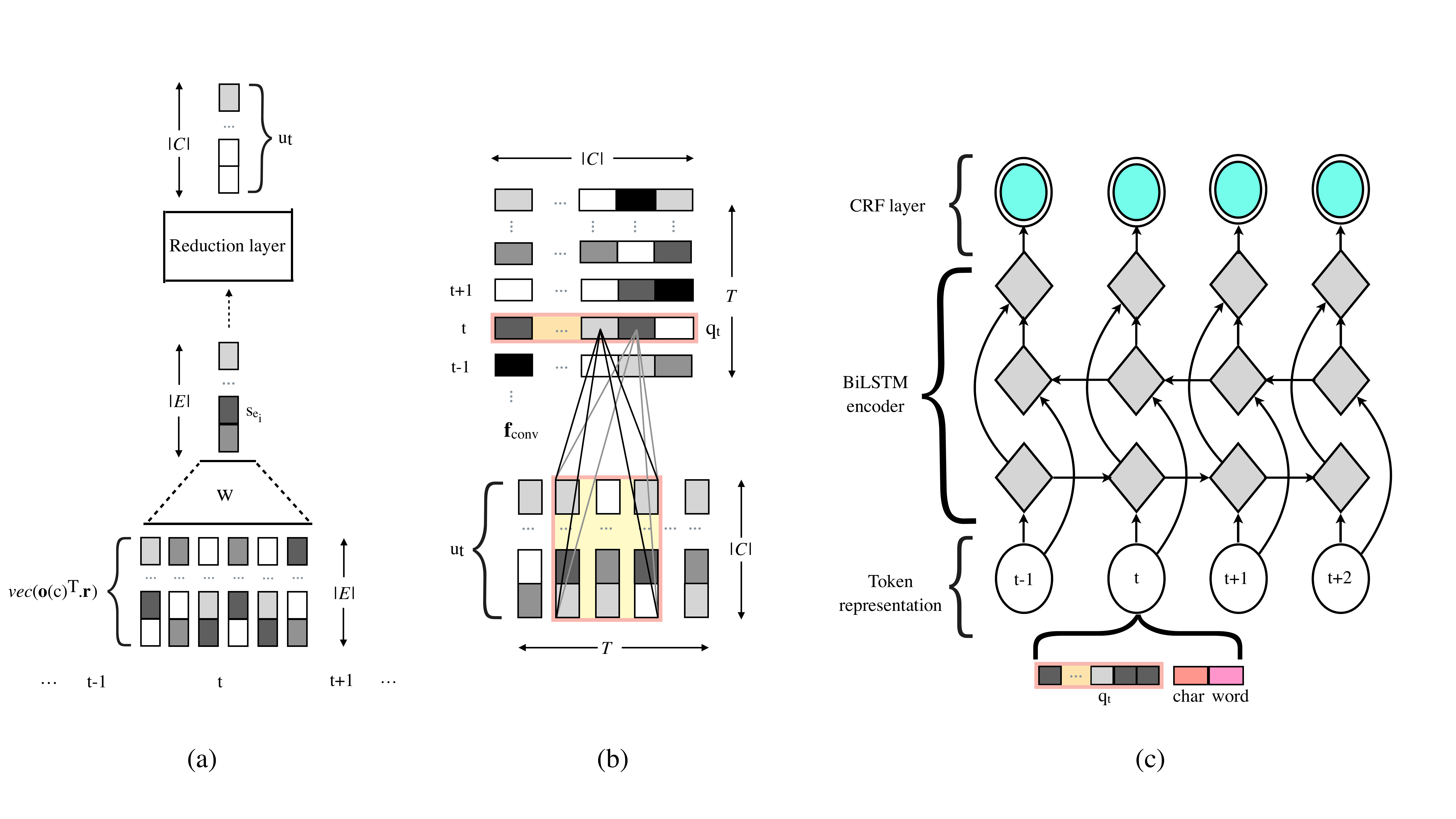}
  \vspace{-0.37in}
  \caption{DEXTER Architecture. (a) the dense aggregation layer (Section \ref{subsec:dense_aggregation}: Dense Aggregation Layer),
  (b) the CNN architecture to capture contextual signals (Section \ref{subsec:contextual_cnn}: Capturing Contextual Entity Signals),
  and (c) the integration of DEXTER embeddings with an NER task (Section \ref{subsec:nertraining}: NER Training).}
  \label{fig:dense}
\end{figure}
\section{Detailed Shallow Parser Results}
\label{sec:app_sp_table}

\begin{table*}[h]
    \begin{center}
    \caption{\label{tab:sp} Results for DEXTER features used in a shallow semantic parser.
    The values shown in the ``Semantic Parse accuracy'' column represents the semantic parser's \% accuracy rates for the corresponding domain.
    ``Baseline'' is the baseline semantic parser model which uses only word and character embeddings.
    ``Baseline $+$ token-match gazetteers'' is the semantic parser model which uses token matched gazetteers concatenated with the word and character embeddings (similar to \cite{muralidharan2021noise}).
    ``Baseline $+$ DEXTER'' is the semantic parser model which uses DEXTER (instead of token matched gazetteers) concatenated with the word and character embeddings.
    \text{*} indicates that the results are statistically significant improvements over the best of the other 2 runs with a $p$-value $<$ 0.05.}
    \begin{tabular}{|l|c|c|c|}
    \hline
    \multirow{2}{*}{Domain} & \multicolumn{3}{c|}{Semantic Parse Accuracy (\%)}\\ \cline{2-4}
    & Baseline & Baseline $+$ token-match gazetteers & Baseline $+$ DEXTER \\ \cline{2-4}
    \hline
    Apps & 90.99 & 90.57 & \textbf{91.49}*{ (0.009)} \\
    Businesses \& POIs & 78.54 & 79.52 & \textbf{80.84}*{ (1.1e-13)} \\
    Movies & 68.6 & 69.83 & \textbf{71.17}*{ (7.3e-6)} \\
    Locations & 82.5 & 83.09 & \textbf{83.94}*{ (2.3e-6)} \\
    Restaurants & 85.19 & 86.39 & \textbf{87.16}*{ (4.9e-4)} \\
    \hline
    Reminders & 85.74 & 85.77 & \textbf{86.27}{ (0.142)} \\
    Notebook & 74.1 & 74.94 & \textbf{75.19}{ (0.298)} \\
    Photos & 75.25 & 75.48 & \textbf{75.69}{ (0.291)} \\
    Clock & 94.28 & 94.37 & \textbf{94.40}{ (0.442)} \\
    Ride & \textbf{90.35}{ (0.416)} & 90.33 & 90.28 \\
    Weather & 94.29 & \textbf{94.55}{ (0.474)} & 94.43 \\
    \hline
    \end{tabular}
    \end{center}
    \vspace{-0.35in}
\end{table*}

\end{document}